\documentclass[conference,letterpaper]{IEEEtran}

\usepackage[utf8]{inputenc} 
\usepackage[T1]{fontenc}
\usepackage{url}
\usepackage{ifthen}
\usepackage{cite}
\usepackage{amsfonts}
\usepackage{amssymb}
\usepackage{graphics}
\usepackage{multirow}
\usepackage{epsfig}
\usepackage{graphicx}
\usepackage{subcaption}
\usepackage{wrapfig}
\usepackage{setspace}
\usepackage{centernot}
\usepackage{mdframed}
\usepackage{acronym}
\usepackage{color,colortbl}
\usepackage{enumitem}

\usepackage{times}
\usepackage[colorinlistoftodos]{todonotes}
\usepackage{comment}
\usepackage[ampersand]{easylist}
\usepackage{pgfplots}
\usepackage[cmex10]{amsmath}

\IEEEoverridecommandlockouts
\begin{document}
\title{GANs with Variational Entropy Regularizers:\\Applications in Mitigating the Mode-Collapse Issue}

 \author{
 Pirazh Khorramshahi$^*$\thanks{$^*$The first two authors equally contributed to this work.}, Hossein Souri$^*$, Rama Chellappa, and Soheil Feizi\\
 University of Maryland, College Park\\
 {\{pirazhkh, hsouri, rama\}@umiacs.umd.edu, sfeizi@cs.umd.edu}
 }

\maketitle

\begin{abstract}
Building on the success of deep learning, Generative Adversarial Networks (GANs) provide a modern approach to learn a probability distribution from observed samples. GANs are often formulated as a zero-sum game between two sets of functions; the generator and the discriminator. Although GANs have shown great potentials in learning complex distributions such as images, they often suffer from the mode collapse issue where the generator fails to capture all existing modes of the input distribution. As a consequence, the diversity of generated samples is lower than that of the observed ones. To tackle this issue, we take an information-theoretic approach and maximize a variational lower bound on the entropy of the generated samples to increase their diversity. We call this approach GANs with Variational Entropy Regularizers (GAN+VER). Existing remedies for the mode collapse issue in GANs can be easily coupled with our proposed variational entropy regularization. Through extensive experimentation on standard benchmark datasets, we show all the existing evaluation metrics highlighting difference of real and generated samples are significantly improved with GAN+VER.  

\end{abstract}


\section{Introduction}
\label{sec:intro}


 Generative Adversarial Networks (GANs) \cite{goodfellow2014generative} have shown great promise in generating realistic samples. They are shown to have variety of applications in computer vision \cite{zhu2017unpaired,radford2015unsupervised, NIPS2016_6194, isola2017image, karras2017progressive,karras2019style,lau2019atfacegan}, natural language processing \cite{yu2017seqgan, fedus2018maskgan, yang2017improving, wang2018sentigan}, semantic segmentation \cite{luc2016semantic, dong2017semantic}, and cybersecurity \cite{hitaj2019passgan, hu2017generating, shi2017ssgan}. GANs are able to generate high quality synthetic text, audio, image, and video which is hardly distinguishable from real data. More specifically, in computer vision, GANs have been widely used for image generation \cite{karras2019style, brock2018large, wang2018high}, super-resolution \cite{ledig2017photo}, restoration \cite{lu2019unsupervised, lau2019atfacegan}, and translation \cite{isola2017image}. 

 \begin{figure}[pt!]
     \begin{subfigure}{0.2\textwidth}
        \includegraphics[width=\textwidth]{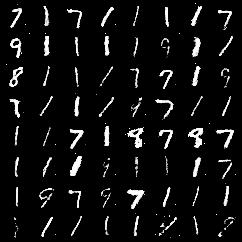}
        \caption{Random Samples of generated MNITS dataset (Vanilla GAN)}
        \label{fig:VGAN_digits}
     \end{subfigure}
     ~
     \centering
     \begin{subfigure}{0.2\textwidth}
        \includegraphics[width=\textwidth]{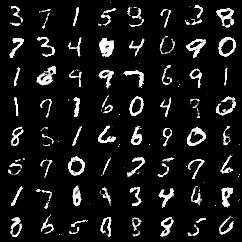}
        \caption{Random samples of generated MNIST dataset (Vanilla GAN+VER)}
     \end{subfigure}
     \caption{Mode collapse issue in generated samples. In (a) generator mostly generates $1,7,9$ modes; however, adding variational entropy regularization increases the diversity of generated samples (b).}
     \label{fig:mode-collapse}
 \end{figure}
 
 In spite of GANs success in generating high quality realistic data, there are issues  which can limit their application. Not only it is not trivial to train a generator and discriminator pair, but also it is hard to evaluate their performance \cite{wang2019generative}. Moreover, {\it Mode collapse} is a common issue during training, where the diversity of generated samples becomes smaller than that of the real data. As an example, with the MNIST dataset with 10 modes e.g. $\{0,\dots,9\}$, the generated samples may only capture few of these modes, as demonstrated in Fig. \ref{fig:VGAN_digits}. Mode collapse in GANs can be the result of improper formulations of training objective \cite{arjovsky2017wasserstein,arora2017generalization}, low-capacity generator \cite{balaji2019normalized,hoang2018mgan} or weak discriminator functions \cite{Li2017TowardsUT,lin2017pacgan}. 
 Main approaches to mitigate the mode collapse issue include (i) {\it discriminator augmentation} such as PacGAN \cite{lin2017pacgan} where the discriminator is modified to make decisions based on multiple samples of either real or generated distributions, (ii) {\it encoder-based regularization} where an encoder function, mapping from the input space to the latent space, is used to increase the diversity of generated samples (e.g. \cite{donahue2016adversarial,dumoulin2016adversarially,srivastava2017veegan}), and (iii) {\it mixture generators} where generated samples probabilistically come from multiple generator functions \cite{balaji2019normalized,hoang2018mgan}. In \cite{balaji2019normalized}, the latter approach has been used and optimized over mixture proportions to mitigate the collapse of rare modes. Although the proposed method in \cite{balaji2019normalized} decreases the mode collapse issue especially on rare modes, it requires knowledge of the number of modes in the data distribution which may not be available. Also, training a GAN with mixture generators is more difficult than the standard GAN used in practice. Another way to reduce mode collapse is to maximize the entropy of the generator \cite{dieng2019prescribed}. However, maximizing differential entropy of a high dimensional random variable is not easy as it requires integration in a high dimensional space. 

In this paper, we propose to take an {\it information-theoretic} approach and use Mutual Information (MI) between input and output of the generator as an alternative to differential entropy. Our proposed approach is complimentary to existing methods and can be effortlessly coupled with them to further mitigate the mode collapse in generative models.

In summary, our contributions in this paper can be summarized as follows:
\begin{itemize}
    \item We prove that MI of input and output of generator is directly related to the entropy of generated samples and maximizing MI directly increases the diversity of generated samples.  
    \item Through extensive experimentation, we show that our proposed approach reduces the degree of mode collapse across standard benchmarks. 
    \item We establish benchmark results on the best GAN evaluation metrics.
\end{itemize}

\section{Related Work}
\label{related work}

Several approaches for mitigating mode collapse have been proposed \cite{tolstikhin2017adagan,srivastava2017veegan,metz2016unrolled,che2016mode,lin2017pacgan,dieng2019prescribed}. Tolstikhin \emph{et al.} \cite{tolstikhin2017adagan} proposed to train a collection of generators instead of one generator, inspired by boosting techniques. Srivastava \emph{et al.} \cite{srivastava2017veegan} proposed VEEGAN to address this problem by employing a reconstructor network which reverses the generator role and maps the generated data to noise. VEEGAN tries to train the generator and reconstructor network jointly. Consequently, the generator will be encouraged to generate the entirety of the true data distribution. Lin \emph{et al.} \cite{lin2017pacgan} proposed to modify the discriminator to make decisions based on multiple samples from a same class; therefore, the discriminator will be able to do binary hypothesis testing which penalizes the generator prone to mode collapse. Recently, prescribed Generative Adversarial Networks (PresGAN) have been proposed by  Dieng \emph{et al.} \cite{dieng2019prescribed}. PresGAN mitigate mode collapse by adding the negative of entropy of the generator to the loss function and tries to maximize the entropy by minimizing the loss function. Since the density of the generator is intractable, PresGAN add noise to the output of the density network and use unbiased estimates to compute gradients of the entropy regularization term. In our model, we avoid computing entropy of the generator directly. Instead, our model benefits from using the MI between the latent variable $Z$ and generated sample $\hat{X}$. In section \ref{method} we prove MI in this case is identical to entropy.
\section{Method}
\label{method}
GANs are based on minimizing a distance measure between two distributions. That is:
\begin{align}
\label{eq:gans}
\min_{\mathbf{\theta}_g}~ d\left(\mathbb{P}_X, \mathbb{P}_{\hat{X}}\right),
\end{align}
where $X$ and $\hat{X}=G(Z; \theta_g)$ represent the distributions of real and generated data respectively, with $Z$ being the input latent variable to the generator. Moreover, $d(.,.)$ is a distance measure between two distributions\footnote{Note that in general, $d(.,.)$ may not be a metric.} and $\theta_g$ is the set of parameters of the generator. In the last couple of years, several distance measures have been used in optimization \eqref{eq:gans} such as {\it optimal transport (OT)} measures \cite{arjovsky2017wasserstein,miyato2018spectral},  {\it divergence} measures \cite{radford2015unsupervised,nowozin2016f}, {\it moment-based} measures  \cite{dziugaite2015training,li2015generative}, etc. In particular, an example of the OT distance is {\it Wasserstein} defined as:
\begin{equation}
\label{eq:Wasserstein}
d(\mathbb{P}_X,\mathbb{P}_{\hat{X}}):=\min_{P_{X,\hat{X}}}~ \mathbb{E}\left[\|X-\hat{X}\|^p\right],
\end{equation}
where $p\geq 1$ is the order of the Wasserstein distance. 

The celebrated Shannon entropy \cite{shannon1948mathematical} is a fundamental measure of diversity in distributions. Therefore, forcing the generator to increase the entropy of the generative distribution should directly impact the mode collapse issue. This can be done through entropy regularization in GANs objective:
\begin{align}
\label{opt:gans}
\min_{\mathbf{\theta}_g}~ d\left(\mathbb{P^{}}_X, \mathbb{P}_{\hat{X}}\right)-\lambda h\left(\mathbb{P}_{\hat{X}}\right),
\end{align}
Where $h(.)$ is the {\it differential entropy} defined as $h(\mu)=\mathbb{E}_{x\sim \mu}[-\log (\mu(x))]$ and $\lambda$ is the regularization parameter. Moreover, the negative entropy function is strongly convex, thus adding entropy regularization term to the formulation of a generative model will improve its optimization landscape as well. This phenomenon has been observed in a related entropy regularization for the optimal transport optimization in \cite{cuturi2013sinkhorn}. Note that the entropy function in \cite{cuturi2013sinkhorn} has been applied on the coupling distribution of an optimal transport while our proposal applies the entropy function to the distribution of the generative model to increase its diversity and resolve the mode collapse issue. 

\begin{figure}
     \centering
     \begin{subfigure}{0.5\textwidth}
     \includegraphics[width=\textwidth]{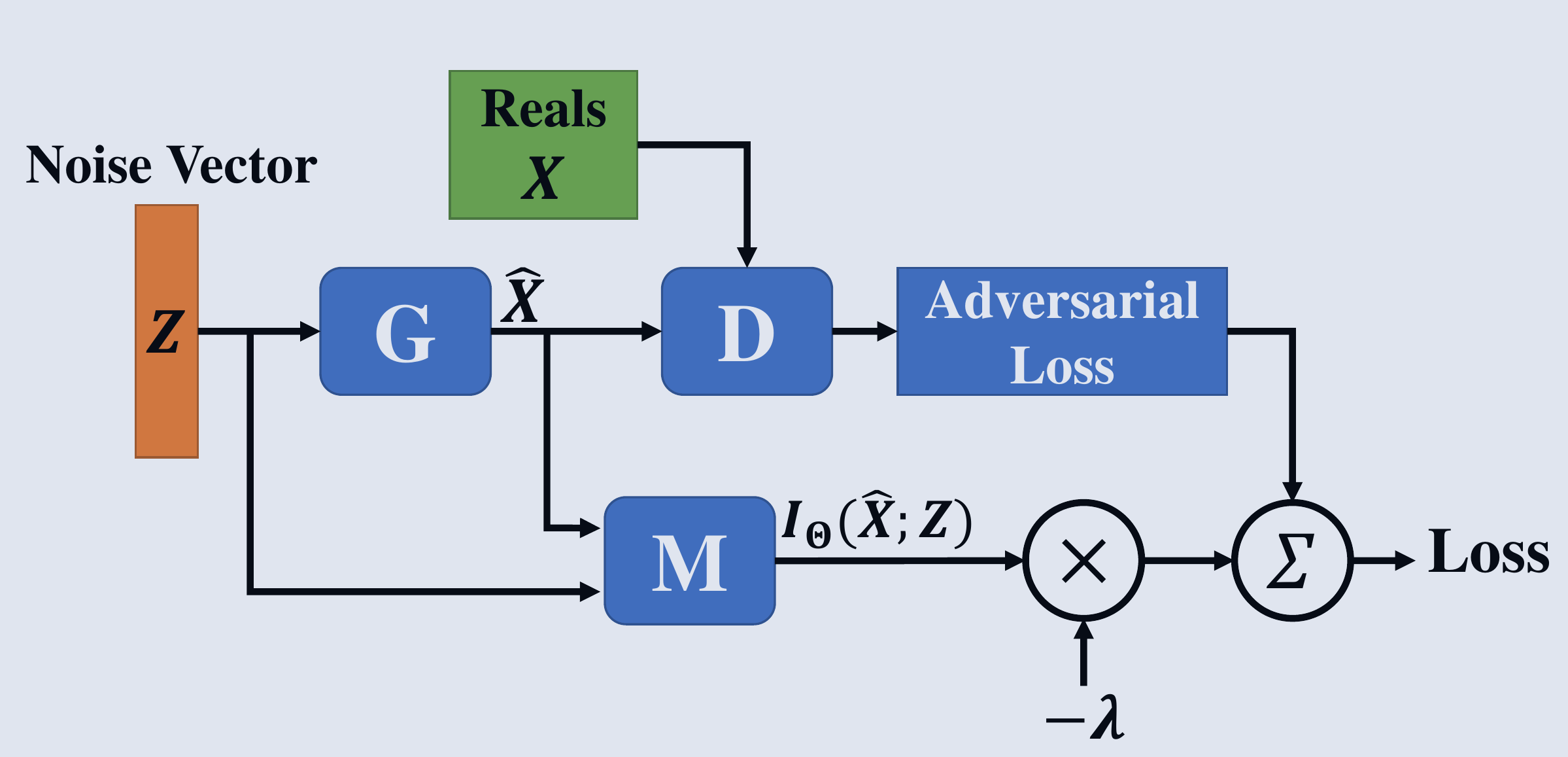}
     \label{subfig:MNIST_digits}
     \end{subfigure}
     \caption{Overview of the proposed model. Noise vector $Z$ is passed to the generator $G$ to generate sample $\hat{X}$. Then real data $X$ and $\hat{X}$ go through the discriminator $D$ to compute adversarial loss. To compute $I_{\Theta}\left(\hat{X}; Z\right)$, $Z$ and $\hat{X}$ are passed through the mutual information estimator network $M$ \cite{belghazi2018mine}. 
     }
     \label{fig:model}
 \end{figure}

Here we consider the mutual information function $I(\hat{X}; Z) = h(\hat{X}) - h(\hat{X}|Z)$. Since the conditional entropy term $h(\hat{X}|Z)$ is zero for a deterministic generator function we have $I(\hat{X}; Z) = h(\hat{X})$ and the optimization problem \eqref{opt:gans} can be written as
\begin{align}\label{opt:gans-information}
\min_{\mathbf{\theta}_g}~ d\left(\mathbb{P}_X, \mathbb{P}_{\hat{X}}\right)-\lambda I\left(\hat{X}; Z\right),
\end{align}
 This optimization is similar to that of the InfoGAN \cite{chen2016infogan} with the difference that InfoGAN introduces extra code variables and maximizes the mutual information between the code and $\hat{X}$ to provide  disentangled representations of the data. However, in our case, we maximize the mutual information between the latent variable $Z$ and generated sample $\hat{X}$ to mitigate the mode collapse issue. To solve the optimization \eqref{opt:gans-information}, we use the neural information measure introduced in \cite{belghazi2018mine} as a variational lower bound on mutual information:
\begin{align}
\label{eq:neural_information_measure}
    I_{\Theta}(\hat{X};Z) = \sup_{\theta_m} \mathbb{E}_{\mathbb{P}_{\hat{X}Z}}\left[T_{\theta_m}\right] - \log\left(\mathbb{E}_{\mathbb{P}_{\hat{X}}\otimes \mathbb{P}_Z}\left[e^{T_{\theta_m}}\right]\right)
\end{align}
where $\mathbb{P}_{\hat{X}Z}$ and $\mathbb{P}_{\hat{X}}\otimes \mathbb{P}_Z$ are joint and product of marginal distributions of $\hat{X}$ and $Z$. Additionally, $T_{\theta_m}: \mathcal{\hat{X}} \times \mathcal{Z} \rightarrow{\mathbb{R}}$ is a family of functions characterized by a deep neural network \cite{belghazi2018mine} that can be optimized using gradient descent methods. Hence, we solve the following optimization problem:
\begin{align}\label{opt:gans-neural_info_measure}
\min_{\mathbf{\theta}_g}\min_{\mathbf{\theta}_m}~ d\left(\mathbb{P}_X, \mathbb{P}_{\hat{X}}\right)-\lambda I_{\Theta}\left(\hat{X}; Z\right)
\end{align}
 Fig. \ref{fig:model} illustrates the block diagram of our proposed model architecture. In the next section, we present our experimental setup, implementation details of the GAN+VER and datasets used in our experiments.
 
\section{Experiments}
In this section we present datasets used in prior works for density estimation and evaluation metrics to assess to quality of generated samples and severity of mode collapse, afterwards we describe the implementation details of the GAN+VER. 

\subsection{Datasets}
Standard benchmarks, namely synthetic Gaussian Mixture Model(GMM) and MNIST have been frequently used to test the effectiveness of various GAN models and we follow the same convention. In the case of GMM we consider $1e5$ random training samples with $8$ mixture components on a ring of radius $1$ and $25$ components on a grid of size $8\times 8$. The standard deviation used in creating GMM datasets is $\sigma=0.05$. During test time, we synthesize $2500$ random sample from real distribution and generate same number of samples from the trained generator for five times and do averaging to compute the evaluation scores.  

\subsection{Evaluation Metrics}
Due to the absence of true target and source distributions, we measure sample-based evaluation metrics designed to assess the quality of generated samples. Therefore, following \cite{xu2018empirical}, for MNIST dataset we use: 
\begin{itemize}
    \item \textbf{Inception Score (IS)}: is the most commonly used metric to evaluate the quality of generated samples using an ImageNet \cite{deng2009imagenet} pre-trained Inception model $\mathcal{M}$ \cite{szegedy2016rethinking}. IS is computed as follows:
    \begin{equation*}
        \text{IS} = e^{\mathbb{E}_{x\sim P_{G}}\left[KL\left(p_{\mathcal{_M}}(y|x)||p_{\mathcal{_M}}(y)\right)\right]}
    \end{equation*}
    where $KL$ is the Kullback-Leibler divergence and $p_{\mathcal{_M}}(y|x)$ and $p_{\mathcal{_M}}(y)$ are the conditional and marginal distribution of Inception model predicted labels. Larger values of IS corresponds to higher quality of generated samples; However, IS score is only meaningful for colored image generation as it is based on ImageNet. Therefore, we train model $\mathcal{M}$ on training set of MNIST dataset and used the trained model in IS calculation. Note that IS is not suited for mode collapse as no comparison between real and generated data occurs in its formulation.
    \item \textbf{Frechet Inception Distance (FID)}: is the second commonly used metric to evaluate the generated samples quality:
    \begin{equation*}
        \text{FID} = || \mu_{_X} - {\mu_{_{\hat{X}}}} || + tr\left(C_{_X} + {C_{_{\hat{X}}}} - 2{({C_{_X}}{C_{_{\hat{X}}}})}^{1/2}\right)
    \end{equation*}
    In FID, distributions of Inception model $\mathcal{M}$ output for real and generated samples are modeled as multivariate Gaussian distributions, i.e. $P_{_X}\sim\mathcal{N}(\mu_{_X}, {C_{_X}})$ and $P_g\sim\mathcal{N}({\mu_{_{\hat{X}}}}, {C_{_{\hat{X}}}})$ respectively. Intuitively, lower values are desired.
    \item \textbf{Maximum Mean Discrepancy (MMD)}: measures the dissimilarity of real $P_X$ and generated $P_{\hat{X}}$ samples distributions with Gaussian kernel $k$ respectively. 
    \begin{equation*}
        \text{MMD} = \mathbb{E}_{\underset{\large \hat{x},\hat{x}^{'} \sim P_{\hat{X}}}{x,x^{'} \sim P_X}}\left[k(x,x^{'}) -2k(x, \hat{x}) + k(\hat{x},\hat{x}^{'})\right]
    \end{equation*}
    Lower values of MMD are preferred.
    \item \textbf{Wassertein Distance (WD)} is defined as:
    \begin{equation*}
        \text{WD} = \underset{P_{X, \hat{X}}}{\inf} \mathbb{E}\left[{\|X-\hat{X}\|}_{_2}\right].
    \end{equation*}
    Generative distribution which is more similar to real distribution results in a lower WD score. 
    
    \item \textbf{1-Nearest Neighbor Classifier (1-NN) Leave Out One (LOO) Accuracy} is used to measure the performance of classifier, i.e., Total Accuracy (TA), class accuracy (Real Accuracy (RA), Generated Accuracy (GA)), Precision (PR) and Recall (RE) in a two class setting. In case that generated samples successfully replicate the properties of real data distribution, all the accuracies, precision and recall should be around $50\%$. According to \cite{xu2018empirical} this metric can best capture the quality of samples in terms of mode collapse. 
\end{itemize}
In the case of GMM datasets in addition to MMD, WD and 1-NN LOO accuracies, we adopt the metrics suggested in \cite{srivastava2017veegan}:
\begin{itemize}
    \item \textbf{Modes}: is the average number of detected mixture components. A component is detected if there is a generated sample which lies within the three standard deviation of it. The issue with this metric is that even if there is only one generated sample close to a component even randomly, it is deemed detected. 
    \item \textbf{High Quality (HQ) Samples}: is the ratio of generated samples which lie within three standard deviation of mixture components. This metric is not indicative of sample's quality as in case of generator's mode collapse it yields $100\%$. 
    \item \textbf{KL Divergence} of the synthetic and generated data. In our implementation we segment the 2D plain into squared regions with sides equal to the standard deviation of mixture models to create bins and measure densities.
\end{itemize}

\begin{figure*}[t!]
    \centering
    \begin{subfigure}[b]{0.2\textwidth}
        \includegraphics[width=\textwidth]{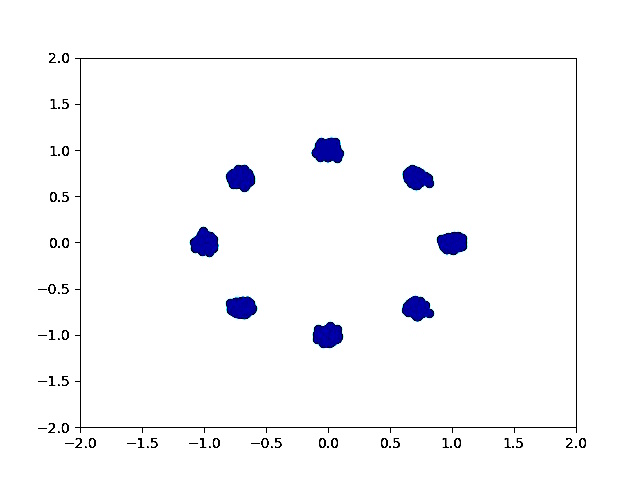}
        \caption{GMM with $8$ modes}
        \label{fig:ring}
    \end{subfigure}
    \hspace{-0.6cm}
    ~
    \begin{subfigure}[b]{0.2\textwidth}
        \includegraphics[width=\textwidth]{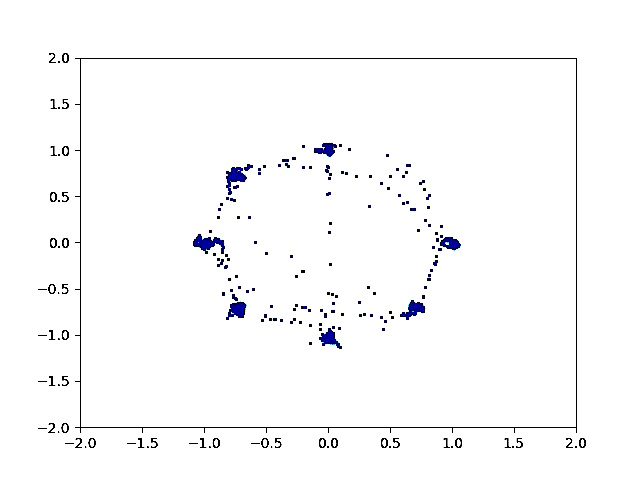}
        \caption{VGAN}
        \label{fig:ring_vgan}
    \end{subfigure}
    \hspace{-0.6cm}
    ~ 
    \begin{subfigure}[b]{0.2\textwidth}
        \includegraphics[width=\textwidth]{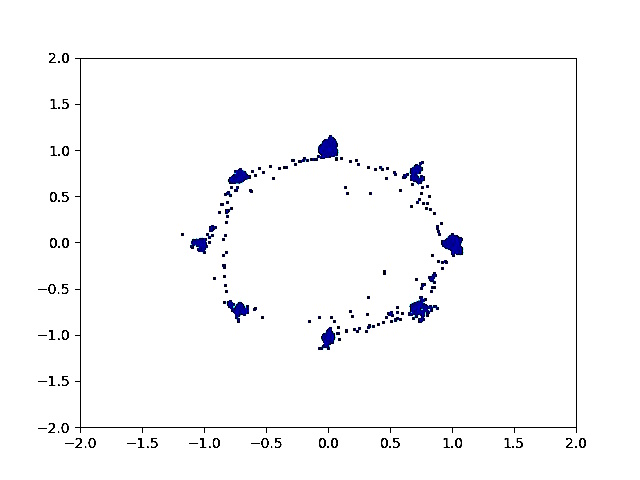}
        \caption{WGAN}
        \label{fig:ring_wgan}
    \end{subfigure}
    \hspace{-0.6cm}
    ~
    \begin{subfigure}[b]{0.2\textwidth}
        \includegraphics[width=\textwidth]{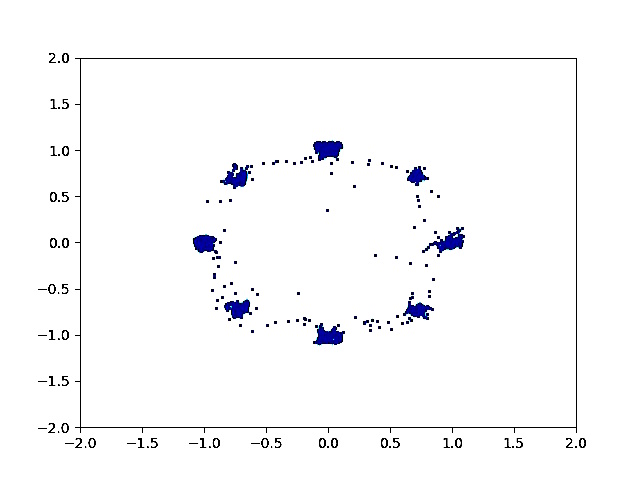}
        \caption{VGAN+VER}
        \label{fig:ring_vgan_MI}
    \end{subfigure}
    \hspace{-0.6cm}
    ~
    \begin{subfigure}[b]{0.2\textwidth}
        \includegraphics[width=\textwidth]{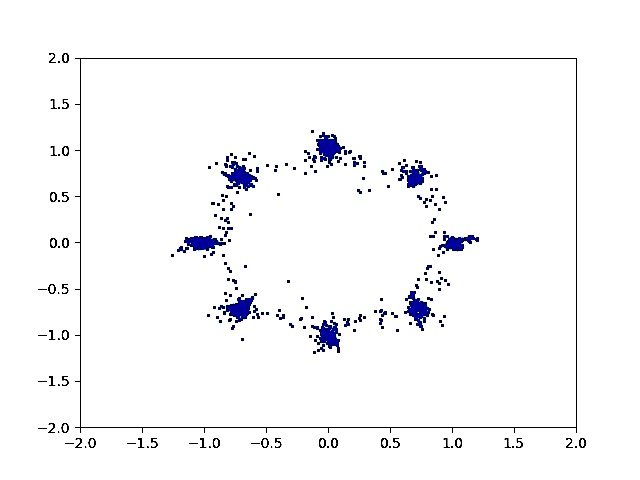}
        \caption{WGAN+VER}
        \label{fig:ring_wgan_MI}
    \end{subfigure} 
    ~
    \begin{subfigure}[b]{0.2\textwidth}
        \includegraphics[width=\textwidth]{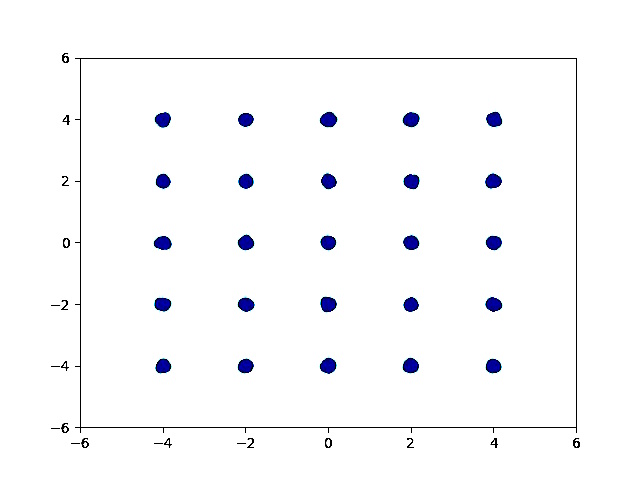}
        \caption{GMM with $25$ modes}
        \label{fig:grid}
    \end{subfigure}
    \hspace{-0.6cm}
    ~
    \begin{subfigure}[b]{0.2\textwidth}
        \includegraphics[width=\textwidth]{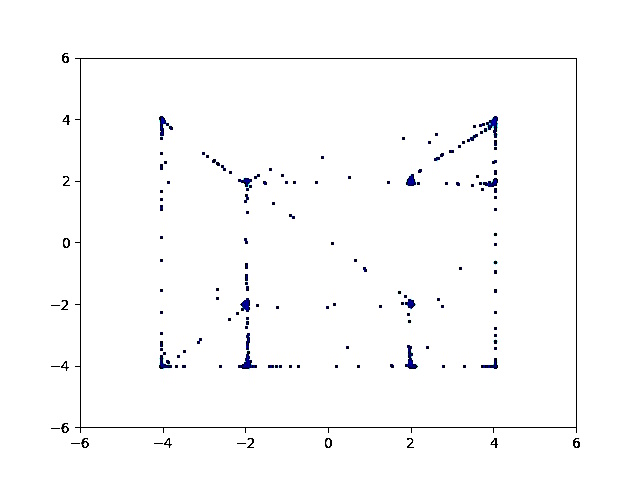}
        \caption{VGAN}
        \label{fig:grid_vgan}
    \end{subfigure}
    \hspace{-0.6cm}
    ~ 
    \begin{subfigure}[b]{0.2\textwidth}
        \includegraphics[width=\textwidth]{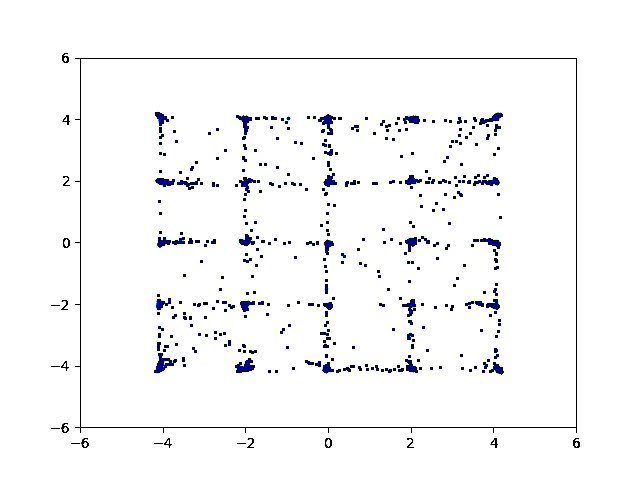}
        \caption{WGAN}
        \label{fig:grid_wgan}
    \end{subfigure}
    \hspace{-0.6cm}
    ~
    \begin{subfigure}[b]{0.2\textwidth}
        \includegraphics[width=\textwidth]{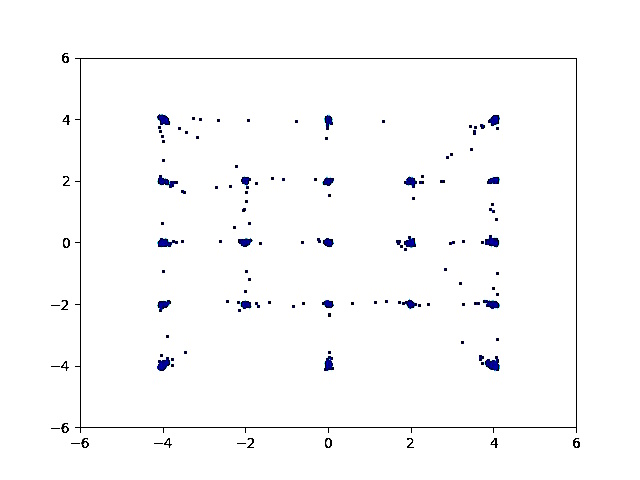}
        \caption{VGAN+VER}
        \label{fig:grid_vgan_MI}
    \end{subfigure}
    \hspace{-0.6cm}
    ~
    \begin{subfigure}[b]{0.2\textwidth}
        \includegraphics[width=\textwidth]{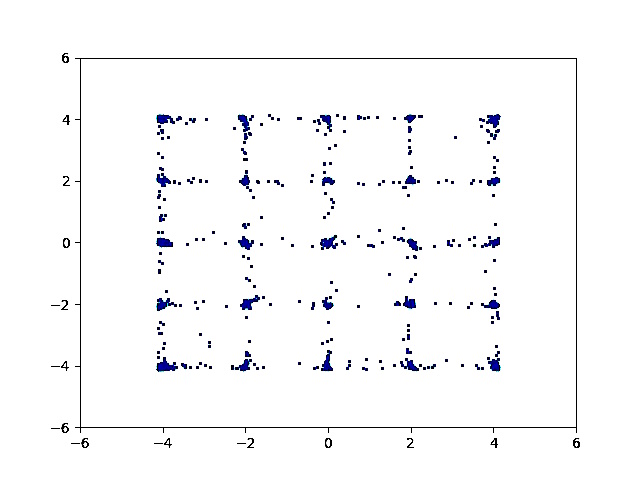}
        \caption{WGAN+VER}
        \label{fig:grid_wgan_MI}
    \end{subfigure} 
    \caption{Synthetic GMM datasets. First row corresponds to generators trained on data with $8$ components. Seconds row represents generated samples from synthetic grid dataset with $25$ modes.}
    \label{fig:GMM}
\end{figure*}

\subsection{Experimental Setup \& Implementation Details}
\label{subsec:setup}
In our experiments we consider four scenarios:
\begin{itemize}
    \item \textbf{Vanilla GAN (VGAN)}: Training objective only consists of the original GAN loss proposed in \cite{goodfellow2014generative}:
    \small{
    \begin{equation}
        \min_{G} \max_{D} \mathbb{E}_{X\sim P_X}\left[\log(D(X))\right] + \mathbb{E}_{Z \sim P_Z} \left[\log(1-D(G(Z)))\right]
     \label{opt:vanilla_gan_obj}
    \end{equation}
    }
    \item \textbf{Wasserstein GAN (WGAN)}: Training objective is to minimize the Earth Mover's Distance (EMD) coupled with the discriminator's gradient penalty $GP$ term \cite{gulrajani2017improved}:
    \small{\begin{equation}
    \min_{G} \max_{D} \mathbb{E}_{X \sim P_X}[D(X)] - \mathbb{E}_{Z\sim P_Z}[D(G(Z))] + GP
    \label{opt:wgan_obj}
\end{equation}}
    \item \textbf{VGAN+VER}: In addition to adversarial loss in \eqref{opt:vanilla_gan_obj}, training objective incorporates the neural information measure $I_{\Theta}(\hat{X};Z)$ to encourage diversity in generated samples.
    \item \textbf{WGAN+VER}: Training objective of EMD is augmented with neural information measure $I_{\Theta}(\hat{X};Z)$.
\end{itemize}

For generator $G$ and discriminator $D$, we follow the architectures employed in \cite{lin2017pacgan}. To estimate the neural information measure, we use a two-layer perceptron $M$ with hidden unit size of $128$ and Leaky ReLU non-linearity. All the networks are initialized with Kaming method \cite{he2015delving}. In addition, for the case of GAN+VER we used adaptive gradient clipping suggested in \cite{belghazi2018mine} to bound gradients flowing back from the $M$; this is required since there is no upper bound on MI. Moreover, in our experiments we used batch processing of size $64$ and Adam optimizer \cite{kingma2014adam} with learning rate of $2e-4$. All the networks, $G$, $D$ and $M$ for the case of GAN+VER, were trained to convergence.
\section{Results}
In this section we present best evaluation scores of the four models discussed in subsection \ref{subsec:setup} obtained via cross validation. We also highlight how VER alleviates the mode collapse.
\subsection{GMM}

\subsubsection{Ring} Table \ref{tab:GMM_RING} presents the quantitative results of VGAN, WGAN, VGAN+VER and WGAN+VER in generation of GMM with $8$ mixture components. It is evident that except for HQ samples metric which is not an informative metric), models with variational entropy regularization perform better, specially for 1-NN LOO metrics getting closer to $50\%$. This indicates real and generated samples have become more indistinguishable through variational entropy regularization. Also, note that WGAN-based models has the lowest WD value as their objective is to minimize the Wasserstein distance. 

\begin{table*}
    \centering
    \caption{Performance of generative models on GMM ring dataset with $8$ mixture components}
    \begin{tabular}{c|c|c|c|c|c|c|c|c|c|c}
    \cline{2-11}
     & \multicolumn{10}{|c|}{Evaluation Metrics} \\ \cline{1-11}
     \multicolumn{1}{|c|}{\multirow{2}{*}{Models}} & \multicolumn{1}{|c|}{\multirow{2}{*}{Modes}} & \multicolumn{1}{|c|}{\multirow{2}{*}{HQ(\%)}} & \multicolumn{1}{|c|}{\multirow{2}{*}{KL}} & \multicolumn{1}{|c|}{\multirow{2}{*}{WD}} &
     \multicolumn{1}{|c|}{\multirow{2}{*}{MMD}} &  \multicolumn{5}{|c|}{1-NN LOO } \\ \cline{7-11}
     \multicolumn{1}{|c|}{} & \multicolumn{1}{|c|}{} & \multicolumn{1}{|c|}{} & \multicolumn{1}{|c|}{} & \multicolumn{1}{|c|}{} & \multicolumn{1}{|c|}{} & \multicolumn{1}{|c|}{TA(\%)} & \multicolumn{1}{|c|}{RA(\%)} & \multicolumn{1}{|c|}{GA(\%)} & \multicolumn{1}{|c|}{PR(\%)} & \multicolumn{1}{|c|}{RE(\%)} \\ \cline{1-11}
     \multicolumn{1}{|c|}{VGAN} & \multicolumn{1}{|c|}{8} & \multicolumn{1}{|c|}{\textbf{99.28}} & \multicolumn{1}{|c|}{1.739} & \multicolumn{1}{|c|}{0.185} & \multicolumn{1}{|c|}{0.058} & \multicolumn{1}{|c|}{74.15} & \multicolumn{1}{|c|}{73.47} & \multicolumn{1}{|c|}{74.84} & \multicolumn{1}{|c|}{74.73} & \multicolumn{1}{|c|}{73.47} \\ \cline{1-11}
     \multicolumn{1}{|c|}{WGAN} & \multicolumn{1}{|c|}{8} & \multicolumn{1}{|c|}{94.43} & \multicolumn{1}{|c|}{0.791} & \multicolumn{1}{|c|}{0.038} & \multicolumn{1}{|c|}{0.020} & \multicolumn{1}{|c|}{62.68} & \multicolumn{1}{|c|}{62.80} & \multicolumn{1}{|c|}{62.55} & \multicolumn{1}{|c|}{62.64} & \multicolumn{1}{|c|}{62.80} \\ \cline{1-11}
     \multicolumn{1}{|c|}{VGAN+VER} & \multicolumn{1}{|c|}{8} & \multicolumn{1}{|c|}{96.92} & \multicolumn{1}{|c|}{\textbf{0.452}} & \multicolumn{1}{|c|}{0.059} & \multicolumn{1}{|c|}{0.011} & \multicolumn{1}{|c|}{\textbf{58.20}} & \multicolumn{1}{|c|}{\textbf{56.72}} & \multicolumn{1}{|c|}{59.68} & \multicolumn{1}{|c|}{\textbf{59.22}} & \multicolumn{1}{|c|}{\textbf{57.40}} \\ \cline{1-11}
     \multicolumn{1}{|c|}{WGAN+VER} & \multicolumn{1}{|c|}{8} & \multicolumn{1}{|c|}{92.71} & \multicolumn{1}{|c|}{0.462} & \multicolumn{1}{|c|}{\textbf{0.019}} & \multicolumn{1}{|c|}{\textbf{0.002}} & \multicolumn{1}{|c|}{58.36} & \multicolumn{1}{|c|}{57.40} & \multicolumn{1}{|c|}{\textbf{59.32}} & \multicolumn{1}{|c|}{59.23} & \multicolumn{1}{|c|}{58.72} \\ \cline{1-11}
\end{tabular}
    \label{tab:GMM_RING}
\end{table*}

\subsubsection{Grid} 
\begin{figure}
    \centering
    \begin{subfigure}[b]{0.24\textwidth}
        \includegraphics[width=\textwidth, height=0.775\textwidth]{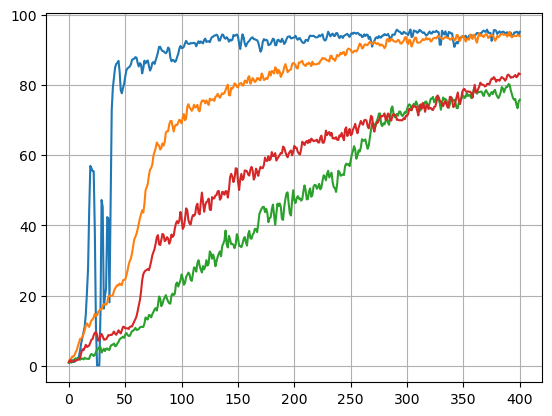}
        \caption{HQ}
    \end{subfigure}
    \hspace{-0.5cm}
    ~
    \begin{subfigure}[b]{0.24\textwidth}
        \includegraphics[width=\textwidth]{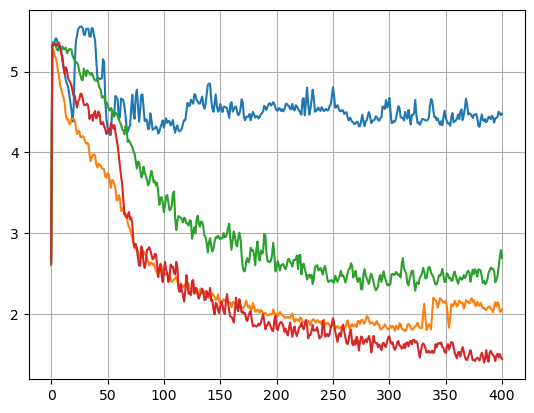}
        \caption{KL}
    \end{subfigure}
    ~
    \begin{subfigure}[b]{0.24\textwidth}
        \includegraphics[width=\textwidth]{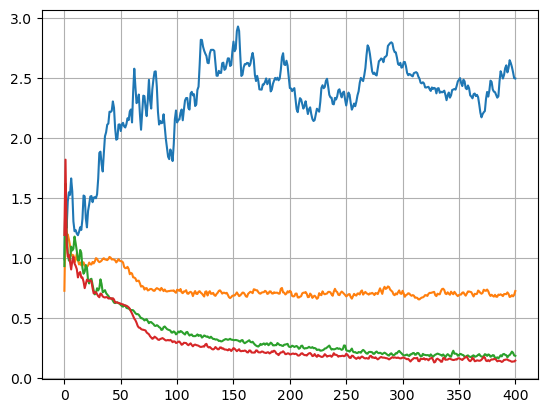}
        \caption{WD}
    \end{subfigure}
    \hspace{-0.5cm}
    ~
    \begin{subfigure}[b]{0.24\textwidth}
        \includegraphics[width=\textwidth]{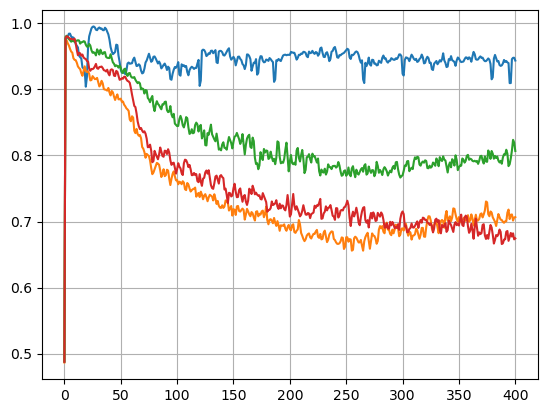}
        \caption{TA}
    \end{subfigure}
    \caption{Progression of evaluation metrics in Table \ref{tab:GMM_GRID} for VGAN (blue), WGAN (green), VGAN+VER (orange) and WGAN+VER (red) when are trained on GMM grid dataset (x-axis represents epochs).}
    \label{fig:GRID_SAMPLE_METRIC}
\end{figure}

Evaluation results for models trained on GMM with $25$ components are reported in Table \ref{tab:GMM_GRID}. Similar to the GMM ring dataset, entropy regularization improves evaluation metrics for both VGAN and WGAN. Also note that due to the better mathematical properties of Wasserstein distance, WGAN's generated samples are superior to VGAN in terms of evaluation metrics as expected, however augmenting the training objective of VGAN with VER boosts its performance close to or even better than WGAN+VER for some 1-NN LOO metrics. Figure \ref{fig:GRID_SAMPLE_METRIC} shows the progression of some metrics for the four models over the course of training. Figure \ref{fig:GMM} visualizes the GMM ring and grid datasets and how successful the four models are in learning the respective densities.

\begin{table*}
    \centering
    \caption{Performance of generative models on GMM grid dataset with $25$ mixture components}
    \begin{tabular}{c|c|c|c|c|c|c|c|c|c|c}
    \cline{2-11}
     & \multicolumn{10}{|c|}{Evaluation Metrics} \\ \cline{1-11}
     \multicolumn{1}{|c|}{\multirow{2}{*}{Models}} & \multicolumn{1}{|c|}{\multirow{2}{*}{Modes}} & \multicolumn{1}{|c|}{\multirow{2}{*}{HQ(\%)}} & \multicolumn{1}{|c|}{\multirow{2}{*}{KL}} & \multicolumn{1}{|c|}{\multirow{2}{*}{WD}} &
     \multicolumn{1}{|c|}{\multirow{2}{*}{MMD}} &  \multicolumn{5}{|c|}{1-NN LOO} \\ \cline{7-11}
     \multicolumn{1}{|c|}{} & \multicolumn{1}{|c|}{} & \multicolumn{1}{|c|}{} & \multicolumn{1}{|c|}{} & \multicolumn{1}{|c|}{} & \multicolumn{1}{|c|}{} & \multicolumn{1}{|c|}{TA(\%)} & \multicolumn{1}{|c|}{RA(\%)} & \multicolumn{1}{|c|}{GA(\%)} & \multicolumn{1}{|c|}{PR(\%)} & \multicolumn{1}{|c|}{RE(\%)} \\ \cline{1-11}
     \multicolumn{1}{|c|}{VGAN} & \multicolumn{1}{|c|}{19} & \multicolumn{1}{|c|}{\textbf{96.43}} & \multicolumn{1}{|c|}{4.160} & \multicolumn{1}{|c|}{1.169} & \multicolumn{1}{|c|}{0.039} & \multicolumn{1}{|c|}{86.87} & \multicolumn{1}{|c|}{90.11} & \multicolumn{1}{|c|}{83.64} & \multicolumn{1}{|c|}{81.85} & \multicolumn{1}{|c|}{90.11} \\ \cline{1-11}
     \multicolumn{1}{|c|}{WGAN} & \multicolumn{1}{|c|}{\textbf{25}} & \multicolumn{1}{|c|}{80.87} & \multicolumn{1}{|c|}{2.21} & \multicolumn{1}{|c|}{0.155} & \multicolumn{1}{|c|}{0.002} & \multicolumn{1}{|c|}{75.54} & \multicolumn{1}{|c|}{75.08} & \multicolumn{1}{|c|}{75.99} & \multicolumn{1}{|c|}{75.93} & \multicolumn{1}{|c|}{75.08} \\ \cline{1-11}
     \multicolumn{1}{|c|}{VGAN+VER} & \multicolumn{1}{|c|}{23} & \multicolumn{1}{|c|}{95.79} & \multicolumn{1}{|c|}{1.764} & \multicolumn{1}{|c|}{0.649} & \multicolumn{1}{|c|}{0.050} & \multicolumn{1}{|c|}{\textbf{64.60}} & \multicolumn{1}{|c|}{\textbf{64.75}} & \multicolumn{1}{|c|}{\textbf{64.44}} & \multicolumn{1}{|c|}{\textbf{64.55}} & \multicolumn{1}{|c|}{\textbf{64.74}} \\ \cline{1-11}
     \multicolumn{1}{|c|}{WGAN+VER} & \multicolumn{1}{|c|}{\textbf{25}} & \multicolumn{1}{|c|}{83.67} & \multicolumn{1}{|c|}{\textbf{1.33}} & \multicolumn{1}{|c|}{\textbf{0.12}} & \multicolumn{1}{|c|}{\textbf{0.002}} & \multicolumn{1}{|c|}{65.50} & \multicolumn{1}{|c|}{65.47} & \multicolumn{1}{|c|}{65.52} & \multicolumn{1}{|c|}{65.50} & \multicolumn{1}{|c|}{65.47} \\ \cline{1-11}
\end{tabular}
    \label{tab:GMM_GRID}
\end{table*}
\begin{table*}
    \centering
    \caption{Performance of generative models on MNIST dataset}
    \resizebox{2\columnwidth}{!}{%
    \begin{tabular}{c|c|c|c|c|c|c|c|c|c|c|c}
    \cline{2-12}
     & \multicolumn{11}{|c|}{Evaluation Metrics} \\ \cline{1-12}
     \multicolumn{1}{|c|}{\multirow{2}{*}{Models}} &
     \multicolumn{1}{|c|}{\multirow{2}{*}{IS}} &
     \multicolumn{1}{|c|}{\multirow{2}{*}{FID}} &
     \multicolumn{1}{|c|}{\multirow{2}{*}{Modes}} &
     \multicolumn{1}{|c|}{\multirow{2}{*}{KL}} & \multicolumn{1}{|c|}{\multirow{2}{*}{WD}} &
     \multicolumn{1}{|c|}{\multirow{2}{*}{MMD}} &  \multicolumn{5}{|c|}{1-NN LOO} \\ \cline{8-12}
     \multicolumn{1}{|c|}{} & \multicolumn{1}{|c|}{} & \multicolumn{1}{|c|}{} & \multicolumn{1}{|c|}{} & \multicolumn{1}{|c|}{} & \multicolumn{1}{|c|}{} & \multicolumn{1}{|c|}{} &  \multicolumn{1}{|c|}{TA(\%)} & \multicolumn{1}{|c|}{RA(\%)} & \multicolumn{1}{|c|}{GA(\%)} & \multicolumn{1}{|c|}{PR(\%)} & \multicolumn{1}{|c|}{RE(\%)} \\ \cline{1-12}
     \multicolumn{1}{|c|}{VGAN} & \multicolumn{1}{|c|}{4.287} & 
     \multicolumn{1}{|c|}{0.034} & \multicolumn{1}{|c|}{7.1} &
     \multicolumn{1}{|c|}{2.691} & \multicolumn{1}{|c|}{0.770} & \multicolumn{1}{|c|}{0.111} & \multicolumn{1}{|c|}{85.19} & \multicolumn{1}{|c|}{80.00} & \multicolumn{1}{|c|}{90.39} & \multicolumn{1}{|c|}{89.28} & \multicolumn{1}{|c|}{80.01} \\ \cline{1-12}
     \multicolumn{1}{|c|}{WGAN} & \multicolumn{1}{|c|}{9.223} & 
     \multicolumn{1}{|c|}{0.012} & \multicolumn{1}{|c|}{\textbf{10}} &
     \multicolumn{1}{|c|}{0.007} & \multicolumn{1}{|c|}{0.025} & \multicolumn{1}{|c|}{\textbf{0.045}} & \multicolumn{1}{|c|}{68.87} & \multicolumn{1}{|c|}{68.09} & \multicolumn{1}{|c|}{69.65} & \multicolumn{1}{|c|}{\textbf{69.15}} & \multicolumn{1}{|c|}{66.39} \\ \cline{1-12}
     \multicolumn{1}{|c|}{VGAN+VER} & \multicolumn{1}{|c|}{8.431} & 
     \multicolumn{1}{|c|}{0.025} & \multicolumn{1}{|c|}{\textbf{10}} &
     \multicolumn{1}{|c|}{0.256} & \multicolumn{1}{|c|}{0.031} & \multicolumn{1}{|c|}{0.069} & \multicolumn{1}{|c|}{74.47} & \multicolumn{1}{|c|}{73.84} & \multicolumn{1}{|c|}{75.09} & \multicolumn{1}{|c|}{74.78} & \multicolumn{1}{|c|}{73.84} \\ \cline{1-12}
     \multicolumn{1}{|c|}{WGAN+VER} & \multicolumn{1}{|c|}{\textbf{9.235}} & 
     \multicolumn{1}{|c|}{\textbf{0.011}} & \multicolumn{1}{|c|}{\textbf{10}} &
    \multicolumn{1}{|c|}{\textbf{0.007}} & \multicolumn{1}{|c|}{\textbf{0.025}} & \multicolumn{1}{|c|}{0.047} & \multicolumn{1}{|c|}{\textbf{68.19}} & \multicolumn{1}{|c|}{\textbf{67.94}} & \multicolumn{1}{|c|}{\textbf{69.45}} & \multicolumn{1}{|c|}{69.93} & \multicolumn{1}{|c|}{\textbf{65.81}} \\ \cline{1-12}
\end{tabular}
}
    \label{tab:MNIST}
\end{table*}

\subsection{MNIST}
For the MNIST dataset, in addition to metrics adopted in GMM datasets except for HQ, we compute IS and FID scores. Note that here Modes metric is the number of MNIST classes that model $\mathcal{M}$ recognized in a set of randomly $2500$ generated images. Table \ref{tab:MNIST} represents the evaluation results of the four models. It can be observed VGAN has inferior performance with respect to other models; specially the average number of modes detected is $7.1$ which illustrates mode collapse issue associated with VGAN; however when its training objective is augmented with variationl entropy regularization, all the metrics are significantly boosted. Figure \ref{fig:mode-collapse} qualitatively shows how VER reduces mode collapse in VGAN. Also thanks to superior properties of Wasserstein loss, entropy regularization does not seem to improve the metrics considerably. This can be a result of increased complexity of the MNIST data manifold compared to GMM synthetic datasets.

\section{Conclusion}
In this paper, an information-theoretic approach is presented to encourage diversity in the generated samples of a GAN model and reduce the mode collapse issue. We present Variational Entropy Regularization (VER) which tries to maximize a variational lower bound on mutual information between input and output of the generator. Theoretically, it is proved this mutual information is identical to the entropy of generated samples. Therefore, this maximization alleviates the mode collapse as supported in our experiments. Through extensive experimentation, it is showed that VER improves metrics corresponding to generated sample quality and indicative of mode collapse issue across standard datasets; we also established the benchmark results on the most informative GAN metrics for MNIST and GMM datasets. Note that VER is a simple regularization term that can be easily added to any type of GAN model.

In the future, we plan to extend variational entropy regularization to other datasets such as CIFAR-10 and CIFAR-100. Further, we will study the impact of VER on the feature level of generator and discriminator to reduce the gap of neural information measure and mutual information to enjoy a better variational bound.


%

\bibliographystyle{IEEEtran}
\bibliography{egbib}

\end{document}